\documentclass[letterpaper]{article}
\usepackage{aaai}
\usepackage{times}
\usepackage{helvet}
\usepackage{courier}

\usepackage{ctable}

\usepackage{mathtools}

\newcommand{\ve}[1]{\mathbf{#1}}
\newcommand{\ma}[1]{\mathrm{#1}}

\begin{document}
%

\title{EmpTransfo: A Multi-head Transformer Architecture for Creating Empathetic Dialog Systems}
\author{Rohola Zandie and Mohammad H. Mahoor\\
Department of Electrical and Computer Engineering, University of Denver, USA\\
rohola.zandie@du.edu and mmahoor@du.edu
}
\maketitle
\begin{abstract}
\begin{quote}
Understanding emotions and responding accordingly is one of the biggest challenges of dialog systems. This paper presents EmpTransfo, a multi-head Transformer architecture for creating an empathetic dialog system. EmpTransfo utilizes state-of-the-art pre-trained models (e.g., OpenAI-GPT) for language generation, though models with different sizes can be used. We show that utilizing the history of emotions and other metadata can improve the quality of generated conversations by the dialog system. Our experimental results using a challenging language corpus show that the proposed approach outperforms other models in terms of Hit@1 and PPL (Perplexity).
 
\end{quote}
\end{abstract}

\section{Introduction}
\noindent Humans have the unique capability to communicate with nuanced emotions through natural languages. Most of the existing conversational dialog systems focus on language understanding and improving the generated responses. Although these features are essential in building dialog systems, they lack empathetic features for conversation, which is essential for quality communication. To increase user' satisfaction, dialog systems need to understand and incorporate emotions to respond with proper emotions. The positive effects of using emotional dialog systems also have been proved in many areas like customer satisfaction and healthcare applications \cite{dino2019}. 

Recent advances in Natural Language Processing (NLP) with the idea of using pre-trained models have led to remarkable results in different NLP tasks.
Even though applying the same idea in dialog systems has resulted in improved models in terms of language understanding and generation, integrating other information like emotions and context knowledge is still challenging.
To build empathetic conversational agents, machines need to have the ability to recognize and predict emotions based on the history of conversations and use them in interacting with users.

Corpora used in building most of the traditional dialog system include general conversations, 

although these datasets are usually large scale, they lack specificity and do not contain metadata such as emotions, topics, personality, etc. Training a system based on general corpora leads to conversational agents that do not understand emotions, lack any personality and tend to produce generic responses such as: \textit{``I don't know.''}. For these reasons, there has been an effort to create higher quality datasets with more contextual information.
For example, the  \textsc{DailyDialog}~\cite{li2017dailydialog} dataset contains information about emotion, topic, and actions. 

\begin{figure}[]
	\centering
	\includegraphics[width=8cm,clip]{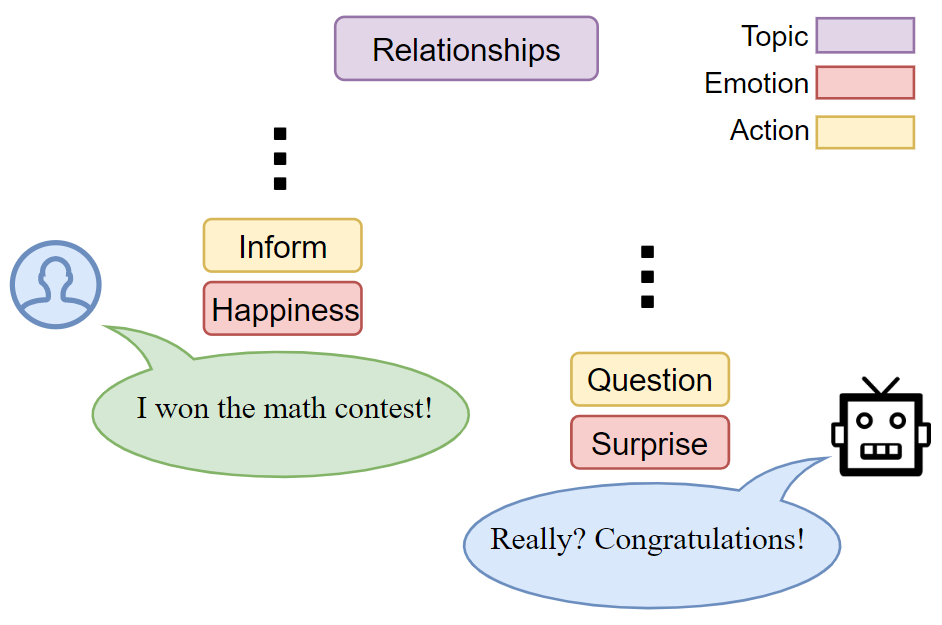}
	\vspace{-3mm}
	\caption{An example of the interaction of EmpTransfo with the user. Contextual information like the history of emotions and actions and also the topic of conversation are crucial to respond with the appropriate emotion. }
	\label{fig:schema}
	\vspace{-1.5em}
	
\end{figure}

We present a novel multi-head Transformer architecture that can use explicit contextual information on emotions, topic and actions to respond to users' utterances with proper emotions without sacrificing the quality of responses in terms of coherence, relevance, and consistency. Figure~\ref{fig:schema} demonstrates how the interaction between the user and dialog system is conditioned on the history of emotions, actions, and the topic. All these contextual clues make it easier for the dialog system to respond with appropriate emotion.  
EmpTransfo is built upon the state-of-the-art dialog system~\cite{wolf2019transfertransfo} and introduces a new architectural design that can exploit contextual information. Quantitative analysis shows the model outperforms all the baseline models. Our main contributions in this paper are:
\begin{enumerate}
\item We incorporate emotions with a multi-task learning approach in dialog systems that is effective and extendable. 
\item We show that our approach can be augmented with other contextual information that not only improves empathetic aspects of responses but also its generation quality. 
\item We design the model in a way that can be used with larger or smaller pre-trained models without changing the architecture of the system. This gives us the flexibility to use EmpTransfo in different settings based on our needs.
\end{enumerate}

\section{Related Work}
\label{sec:relatedwork}
Most of the work on utilizing emotions in conversational systems use large Twitter datasets~\cite{sordoni2015neural} that contain emojis as the meta-information for the emotions. In \cite{zhou2017mojitalk}, they use the Twitter dataset and apply a preprocessing method to create a conversational dataset with 64 different emojis that represent different emotions. In the preprocessing step, they used tweets and responses that contain at least one emoji and filtered other emojis based on the occurring frequency. They used a CVAE \cite{sohn2015learning} network to train and generate emotional responses. The choice of using emojis to represent emotion is noisy because it is too fine-grained and in many cases, the combination of different emojis hardly corresponds to any specific emotions. 


In \cite{colombo2019affect}, they used a seq2seq framework with vector representation for emotions, desired emotion, a regularizer to penalize neutral words and a sampling method that forces the generation of emotionally relevant words.

There are two papers on NLPCC dataset \cite{10.1007/978-3-662-45924-9_10}, a Chinese language dataset with eight emotion categories. The first one is Emotional Chatting Machine \cite{zhou2018emotional} that uses a seq2seq architecture with embedding of emotions along with words and an internal and external memory mechanism to generate emotional responses. The second work is EmoDS \cite{song2019generating} that uses a seq2seq approach with a training objective based on an emotion classifier that promotes implicit emotion generation. Both works use the lexical attention mechanism in decoder with more focus on emotional words to inject explicit emotions into responses. 

Based on the reports from above-mentioned works, all the seq2seq based models tend to generate generic responses that can't capture all the emotions equally well.
Furthermore, all previous works use a machine annotating approach in the training process that introduces noise in results. 

The most relevant work to ours is \cite{rashkin2019towards}. They introduced a new dataset called \textsc{EMPATHETICDIALOGUES} that contains meta-information about conversations. The meta-information is a label that shows emotion and also the situation in which the conversation has happened. 
They proposed two architectures, one retrieval-based model which looks for the best match using the BERT encoder \cite{devlin2018bert} and a generative model using Transformer architecture.
Using one label for the whole conversation and not each utterance makes it harder for the models to find proper correlations. 

On the other hand, recent progress has shown promising results on pre-trained language models on conversational models in chit-chat settings. Recently, \cite{wolf2019transfertransfo} showed that using a fine-tuned GPT (Generative Pre-Training), they can beat any other model on the domain of personal chat using the \textsc{PersonaChat} dataset \cite{zhang2018personalizing}. 

This paper proposes a new architecture that can incorporate not only emotion but other relevant meta information in the  \textsc{DailyDialog} dataset. We first discuss how to use multi-task learning for ``next emotion prediction'' besides language modeling and ``next utterance prediction''.

\section{Proposed Approach}
\label{sec:proposed_approach}

Recent developments have shown substantial improvements in benchmarks on a variety of languages understanding tasks through the use of deep pre-trained language models \cite{devlin2018bert}. More specifically, researchers have shown that by fine-tuning a pre-trained model for specific tasks, they can achieve better performance compared to training the model from scratch. This is also crucial when the dataset at hand is small.

Transformer-based models become ubiquitous in NLP with the work of \cite{vaswani2017attention} for multiple tasks including language generation. Causal language models like GPT and GPT-2 produce remarkable results in the language generation task \cite{wolf2019transfertransfo}. In this paper, we use GPT pre-trained models to achieve better results.

\begin{figure}[tb]
                \includegraphics[width=\columnwidth]
                {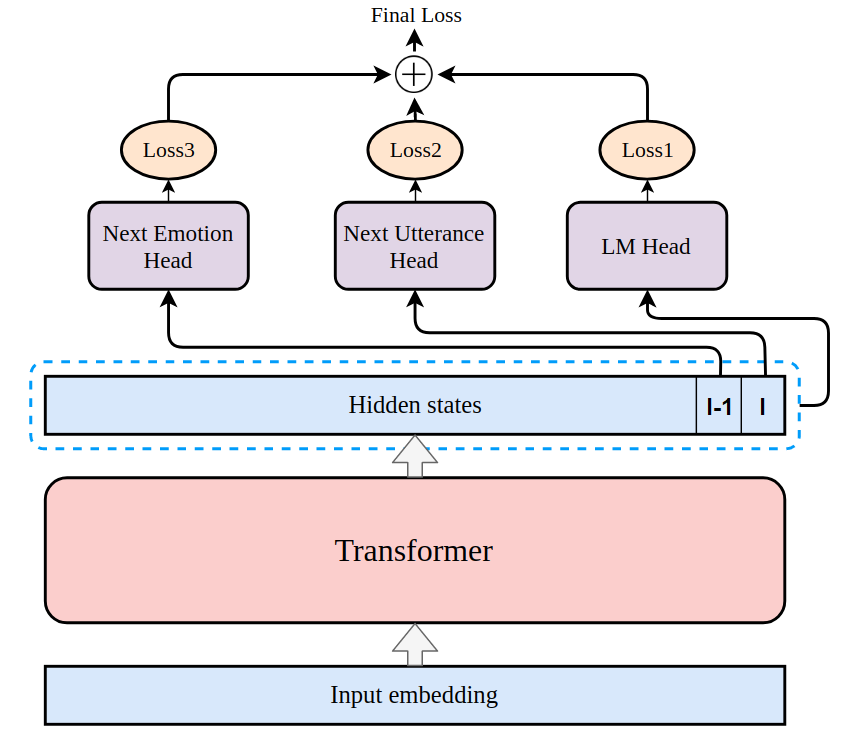}
                \vspace{-2em}
                \caption{EmpTransfo: A multi-head Transformer architecture. There are three feedforward linear heads on top of the Transformer that map different parts of the last layer hidden state to desired output sizes to create the loss functions for language modeling, next utterance prediction, and next emotion prediction. The final loss is a weighted sum of all the losses }
                \label{fig:multi-head_transformer}
                \vspace{-1em}
\end{figure}

\subsection{Empathetic Dialog Generation}
\label{sec:empathetic_dialog_generation}


\begin{figure*}[t]
                \centering
                \includegraphics[width=\textwidth,height=3.5cm]{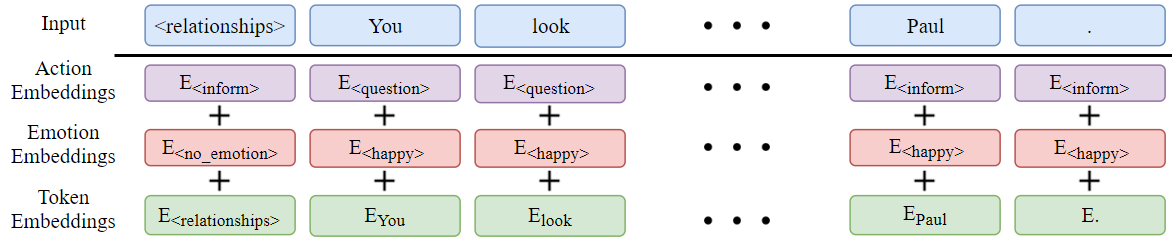}
               
                \caption{The Input representation}
                \label{fig:input_representation}
                
\end{figure*}

Let's assume that in a conversation between two agents, each turn of the conversation by one of the agents is named ``utterance''. Hence, a conversation consists of a set of utterances. More formally, we have \(n\) utterances  \(U=\{u_1, u_2, ..., u_n\}\) and for any utterance \(i\) we have \(N_i\) tokens \(U_i=\{t_1, t_2, ..., t_{N_i}\}\). Also, for each utterance,there is an emotion corresponding to it, resulting in the sequence of emotions \(E=\{e_1, e_2, ..., e_n\}\).

In our dataset, a sample is a sequence of utterances \(\{u_1, u_2, ..., u_{T-1}, u_{next}\}\) in which \(u_{next}\) can be either the correct next utterance \(u_T\) or a distractor from the set of distractors \(U'_{T}\). A distractor is a random utterance from the dataset. In the same way, if the corresponding sequence of emotions is \(\{e_1, e_2, ..., e_{T-1}, e_{next}\}\), then \(e_{next}\) is either the correct next emotion \(e_T\) or a distractor from the set of distractors \(E'_{T}\). A distractor emotion is a random emotion other than \(e_T\) from the set of all emotions.

Our model takes a sequence as input in the embedding space and passes it into a Transformer.
The Transformer architecture \cite{vaswani2017attention} consists of a multi-layer Transformer decoder block. Each Transformer decoder block applies a masked multi-headed self-attention operation followed by a feedforward and a normalization layer over the input hidden states and gives the same size hidden states in the output. We then feed the output of the Transformer to three feed-forward linear heads, responsible for generating the next emotion, utterance, and token. Here, we used a 12-layer architecture but it can be extended or reduced to other model sizes. In the following, we define these three different heads and their corresponding loss functions.

\textbf{1-Language modeling head:} Language modeling is the task of predicting the next token given a sequence of tokens as the context. If we have a sequence of tokens for the correct next utterance as \(U_T=\{t_1, t_2, ..., t_{N}\} \), then the conditional probability of the next token is:

\begin{equation} \label{eq:1}
P(t_i|t_1, ...t_{i-1}) = \text{softmax}(\ve{h}*\ma{W_1})
\end{equation}

In which, \(\ve{h}\) is the last hidden layer of the transformer model and \(\ma{W_1}\) is the token embedding matrix that is learned in training. Then we can define the loss function based on cross-entropy as: 

\begin{equation} \label{eq:2}
\mathcal{L}_1(U_T) = -\sum_{i=1}^{N} \mathrm{log} P(t_i|t_1,...,t_{i-1}) 
\end{equation}
where the context of all previous tokens is encoded in a fixed dimension vector.
It should be noted that the language modeling loss is not trained on the set of next utterance distractors \(U'_T\).

\textbf{2-Next utterance prediction head:} Following \cite{devlin2018bert}, in the next utterance prediction, we try to train the model to predict the next utterance in the conversation. The model learns to distinguish between the correct next utterance among a set of random distractors from other parts of the dataset. More specifically, we create a classifier to calculate the probabilities of the next utterance:
\begin{equation} \label{eq:3}
P_u(a|u_1, u_2, ..u_{T-1})=\text{softmax}(\ve{h_l} * \ma{W_{2}}) 
\end{equation}

\noindent and \(a\) is defined as:

\begin{equation} \label{eq:4}
 a =
  \begin{cases*}
    1 \quad&  $ u_{next} = u_T $ \\
    0 &  $ u_{next} \in U'_T $ \\
  \end{cases*}
\end{equation}

\noindent \(\ve{h_l}\) is the hidden state for the last token from the Transformer decoder and \( \ma{W_{2}}\) is the weight matrix that is learned for the utterance prediction.
Then, the loss function based on cross entropy is:

\begin{equation} \label{eq:5}
    \mathcal{L}_2(U_{1:T}) = -\mathrm{log} P_u(a|u_1, u_2, ..u_{T-1})
\end{equation}

\textbf{3-Next emotion prediction head}: Similar to the next utterance prediction, the model is trained to distinguish between the correct next emotion among a set of distractors. The reason to add this head is to make the model learn not only the grammatical and language structure but also the appropriate emotions for any given history of utterances. We can define:
\begin{equation} \label{eq:6}
  P_e(e| e_1, e_2, ..., e_{T-1}) = \text{softmax}(\ve{h_{l-1}} * \ma{W_3})  
\end{equation}
where \(e\) represents:
\begin{equation}  \label{eq:7}
e =
  \begin{cases*}
    1 \quad&  $ e_{next} = e_T $ \\
    0 &  $ e_{next} \in E'_T $ \\
  \end{cases*}    
\end{equation}
and \(\ve{h_{l-1}}\) is the hidden state of one to the last token from the Transformer decoder and \(\ma{W_3}\) is the weights to be learned during the training for the emotion prediction task. The loss function for next emotion prediction is defined with cross-entropy:

\begin{equation} \label{eq:8}
  \mathcal{L}_3(U_{1:T-1}) = -\mathrm{log} P_e(e|e_1, e_2, ..e_{T-1})  
\end{equation}

Finally, we optimize the following objective which is the total loss function:
\begin{equation} \label{eq:9}
    \mathcal{L}_{total}=c_1\mathcal{L}_{1}+c_2\mathcal{L}_{2}+c_3\mathcal{L}_{3}
\end{equation}
where \(c_1\), \(c_2\), \(c_3\) are hyperparameters that are tuned experimentally. In our experiments, we design the models with and without the ``next emotion prediction'' head for comparison.

\subsection{Input Representation}
\label{sec:input_representation}

We use  \textsc{DailyDialog} \cite{li2017dailydialog} dataset, which is labeled with emotion and action tags per utterance, and with a topic tag for the whole conversation. In Table~\ref{tab:results}, a sample conversation in the preprocessed dataset is shown. A conversation is a sequence of utterances and each utterance can be more than one sentence, but the emotion and action information are defined per utterance. We also add distractors to each sample in the dataset. In the table, the row highlighted in red with the number $d$ shows a distractor.

All the models use learned positional embeddings with a length of up to 512. 
Figure \ref{fig:input_representation} demonstrates the input representation. The embeddings that have been used are:

\begin{enumerate}
\item \textbf{Token embedding:} The input sentences are tokenized using byte pair encoding (BPE) with a vocabulary size of 40,478.

\item \textbf{Emotion embedding:} Each one of seven emotions are considered as a special token to be learned as a new embedding. Emotion embeddings are copied for each token in the utterances and are added to the input of the network.

\item \textbf{Action embedding:} There are four actions for different communication functions that are used in dialog. The dialog acts are: Inform, Question, Directives, and Commissive. Dialog acts are also embedded with special tokens. 

\item \textbf{Topics:} There are 10 topics defined in \textsc{DailyDialog} that are specified for each conversation. For topics, we just concatenate topic embeddings to the beginning of the first input token embedding.

\end{enumerate}
We sum all the embeddings and then feed them to the network.

\begin{table}[]
\label{tab:dataset_sample}
\caption{A conversation in DailyDialog dataset}
\resizebox{\columnwidth}{!}{
\begin{tabular}{llll}
\hline
\textbf{\#} & \textbf{Utterance}                                 & \textbf{Emotion} & \textbf{Action} \\ \hline
1           & You look so happy, any good news?                  & happiness        & question        \\ \hline
2           & Yes, I've won the math contest & happiness        & inform          \\ \hline
3           & Really? Congratulations!                           & surprise         & question        \\ \hline
4           & Thank you Paul.                                    & happiness        & inform          \\
\textcolor{red}{d}           & \textcolor{red}{I really want to take him on my knee.}              & \textcolor{red}{anger}            & \textcolor{red}{inform}         
\end{tabular}
}
\vspace{-1.0em}
\end{table}


\section{Training}
\label{sec:training}

We used the OpenAI pre-trained model on BookCorpus dataset \cite{zhu2015aligning} which covers more than 7,000 books. 
The books include narratives and dialogues, and emotions in a wide range of interactions between characters. This makes the pre-training suitable for the task of dialogue system training because it consists of sentences in a logical order without shuffling. 

Starting with pre-trained weights, we fine-tune our model on the  \textsc{DailyDialog} dataset with the features mentioned in the Input Representation section~\ref{sec:input_representation}. We use the combination of public evaluation and test as the validation set. After preprocessing the training set size is 76,502 and the validation size is 13,809. 

We modify the dataset representation to cover different window positions of conversation history. Each sample in the modified dataset consists of the topic, last two utterances as history context, and the target utterance that can be either the real target or the distractor. The input window then moved forward to cover other parts of the conversation.


We fine-tuned the model with a batch size of 4 for a sequence length of 310 with 20 epochs over the training set of \textsc{DailyDialog} dataset, this is about 1,500,000 steps. For the optimization of the loss function, we used Adam optimizer with a learning rate of 6.25e-5, \(\beta_1=0.9\) and \(\beta_2=0.999\) that decays linearly. The gradient accumulation step is set to 8 with a clipping gradient norm of 1. 
We set the loss coefficients equal to one (\(c_1=c_2=c_3=1\)). The dropout rates for Transformer were borrowed from OpenAI GPT \cite{radford2018improving}. All the proposed models are implemented in Pytorch using Transformers library \cite{Wolf2019HuggingFacesTS}  \footnote{https://github.com/roholazandie/EmpTransfo}.

Here, we use the nucleus top-p sampling for language decoding \cite{holtzman2019curious}. Given logits \(u\) of the last hidden layer, and a sequence of \(i-1\) tokens, \(t_{1:i-1}\) as context, we have the following distribution over the next token \(t_i\) : 

\begin{equation} \label{eq:10}
   P(t_i=V_l | t_{1:i-1}) =\frac{\text{exp}(u_l/T)}{\sum_{l'} \text{exp}(u_{l'}/T)}  
\end{equation}
In which \(V_l\) is the $l^{th}$ token in the vocabulary and \(T\) is the temperature parameter. Higher values of \(T\) result in more stochastic choices over tokens, though lower values of \(T\) approach greedy and deterministic choices for the next token. Based on nucleus top-p sampling, we select \(V^{(p)} \subset V\) as the smallest set such that
\begin{equation} \label{eq:11}
    \sum_{V^{(p)} \subset V} P(t_i|t_{1:i-1}) \geq p
\end{equation}

And then the distribution of Equation \ref{eq:10} should be rescaled to form a probability distribution.
According to \cite{holtzman2019curious}, \(p=0.9\) and \(T=0.7\) are closer to human text generation statistics and we use that for all our experiments.


\begin{table}[]
\centering
\caption{Summary of the results on \textsc{DailyDialog} evaluation set.}
\resizebox{\columnwidth}{!}{
\begin{tabular}{lcccc}
\hline
\multicolumn{1}{c}{\textbf{Model}}    & \textbf{Hit@1} $\uparrow$ & \textbf{PPL} $\downarrow$  & \multicolumn{1}{c}{\textbf{F1}} $\uparrow$ & \textbf{BLEU} $\uparrow$\\ \hline
Seq2Seq+Attention                       & 9.41           & 129.3         & 10.22                            & 5.58           \\ \hline
Transformer ranker                      & 17.20          & -             & \textbf{26.37}                   & \textbf{15.79} \\ \hline
OpenAI GPT without emotion              & 75.01          & 10.19         & 18.2                             & 3.755          \\ 
\specialrule{.12em}{.05em}{.05em} 
EmpTransfo                  & 77.25          & 10.63         & 19.39                            & 3.99           \\ \hline
EmpTransfo + topic          & 76.87          & 10.23         & 18.37                            & 4.51           \\ \hline
EmpTransfo + action         & 77.73          & 9.17          & 18.86                            & 3.71           \\ \hline
EmpTransfo + action + topic & \textbf{78.47} & \textbf{9.04} & 17.27                            & 2.45           \\ \hline

\end{tabular}%
}

\label{tab:results}
\vspace{-1em}
\end{table}


\section{Results}
\label{sec:results}
We evaluated our model on its ability to produce coherent and relevant utterances on the evaluation set. We also evaluate the proposed model on the task of generating emotional responses given the context on the evaluation set. Evaluation of the dialog systems can be done using automatic metrics and human evaluations. Here we restrict the evaluations to automatic metrics.

\begin{table*}[]
\caption{Examples of model responses.}
\label{tab:responses1}
\resizebox{\textwidth}{!}{
\begin{tabular}{l*{3}{c}}
\toprule
& \multicolumn{3}{c}{Input Prompt} \\
\cmidrule(lr){2-4}
Model & I finally passed all the exams! & I failed the exam & You scared me!  \\
\midrule
Seq2Seq+Attention & I'm sorry, but I'm not sure. &  I'm going to go to the some time. & I'm going to go to the job \\
Transformer Ranker & How big was it ? & You're telling me! There are thousands of people here. & Come on! It is really a fun game .   \\
OpenAI GPT w/o emotion & you look much better than before. & let me take your place. & what were you doing?  \\

\midrule
 EmpTransfo+action+topic & that's great! you are really a genius. & Maybe you can try harder next time. & i'm so sorry. i thought you were not coming.  \\

\bottomrule
\end{tabular}
}
\end{table*}

\subsection{Evaluation on coherence and relevance}

\textbf{Baseline models}: For comparison, we used a seq2seq model with attention mechanism and a retrieval-based Transformer ranker dialog system as the baseline. The retrieval-based model is similar to \cite{rashkin2019towards},  created in ParlAI framework \cite{miller2017parlai}.

Seq2Seq+Attention model uses linear attention with a hidden size of 128, learning rate of 0.01 and trained for 20 epochs. Transformer ranker uses a hidden size of 300 that is trained in 40 epochs with cross-entropy loss function. (all other hyperparameters are the defaults from the ParlAI repository).

\textbf{Evaluation metrics}: We use four different metrics to evaluate the models:

\begin{enumerate}
\item Hit@1: this metric is the accuracy of retrieving a gold next utterance among 19 random distractor responses sampled from other dialogues in the dataset.

\item Perplexity (PPL): perplexity is a measure of how well a language model predicts next tokens from the evaluation dataset. More specifically, it is the average per-token log probability over the evaluation set:

\begin{equation}
PPL(p) = e^{\frac{-1}{N}\sum_{w_i}Ln(p_{w_i})}    
\end{equation}

\item BLEU: it is a metric to measure the distance between machine-generated text with human golden labels \cite{papineni2002bleu} .

\item F1 token: measures the F1 score for token level comparison between the generated text and the golden labels.
\end{enumerate}

In Table~\ref{tab:results}, we observe that all our proposed EmpTransfo models outperform baseline models in terms of Hit@1 and PPL. With more contextual features Hit@1 and PPL are improved. The proposed model also shows a significant improvement over \cite{shen2018improving} which has a PPL=23.8 over the same dataset. It also outperforms the Transformer retrieval-based model introduced in \cite{rashkin2019towards}. 

Transformer ranker model has a greater F1 and BLEU compared to other models which is expected because those metrics give higher scores for stronger similarity with the golden utterances in the datasets. BLEU is initially developed for machine translation and studies show that it is not a good metric for text generation evaluation \cite{sulem2018bleu,novikova2017we}. Table~\ref{tab:results} also shows that adding more contextual information like topic and action results in higher Hit@1 and lower PPL. The reason behind this observation is that more contextual information provides the model with better information on selecting the correct next sentence and correct next token.

Table~\ref{tab:responses1} shows some responses with different given input prompts. The inputs are selected in a way to expect the model to respond with emotions. All the outputs are the first results obtained from the models.

\begin{figure}[tb]
                \includegraphics[width=.95\columnwidth]{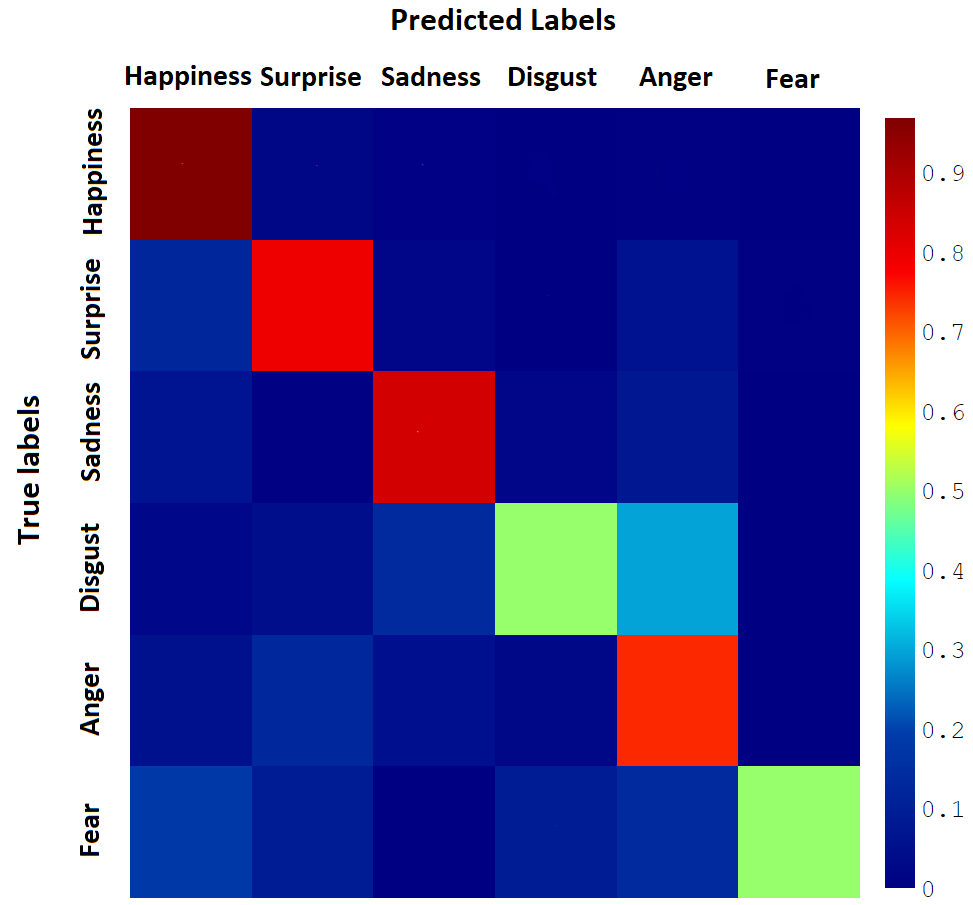}
                \caption{The confusion matrix of emotion prediction for  \textsc{DailyDialog} with 6 emotions using EmpTransfo and all the features (best in color).}
                \label{fig:confusion_matrix}
                \vspace{-1em}
\end{figure}

\subsection{Evaluation on emotion prediction}
In order to evaluate the next utterance emotion prediction, we calculate the precision and recall from the confusion matrix over the evaluation dataset. Figure \ref{fig:confusion_matrix} demonstrates the calculated confusion matrix with Precision=81.35, Recall=72.37 and F1=76.59. The proposed model achieves more than 3 percent improvement compared to \cite{chan2018encoding} whom report their best emotion prediction results with precision=70.81, recall=76.16 and F1=73.39. The confusion matrix also shows interesting observations such as the high rate of confusion between disgust and anger which has been observed in computer vision and facial expression recognition \cite{hasani2017facial}.

\section{Discussion}
\label{sec:discussion}
This paper introduced EmpTransfo a multi-head Transformer model which is an empathetic aware dialog system to interact with users with higher quality in terms of coherence, relevance, and emotion. Our proposed method is built upon the state-of-the-art language model of OpenAI-GPT for language generation.
One of the limitations in the proposed approach is requiring meta information of emotion, action, and topic in order to respond with the proper emotions.

\section{Conclusion and Future Work}
\label{sec:conclusion}

EmpTransfo is a scalable and fully data-driven neural conversational model that effectively exploits the information about emotion, action, and topic. It naturally combines conversational and non-conversational data through multi-task learning. This shows that multi-task learning for training conversational models is not only possible but necessary and can be extended to include more contextual data. As a future direction, we will use knowledge-base and other contextual information to develop knowledge-aware dialog systems.

\bibliography{reference}
\bibliographystyle{aaai}

\end{document}